\documentclass[10pt]{article}

\usepackage{arxiv}
\usepackage[utf8]{inputenc}
\usepackage[T1]{fontenc}
\usepackage{inconsolata}
\usepackage[numbers,sort&compress]{natbib}
\usepackage{graphicx}
\usepackage{float}
\usepackage{booktabs}
\usepackage{array}
\usepackage{amsmath}
\usepackage{amsfonts}
\usepackage{microtype}
\usepackage{authblk}
\usepackage{xcolor}
\usepackage{comment}
\usepackage{url}
\usepackage{hyperref}

\definecolor{arxivlink}{HTML}{1F4E79}
\hypersetup{
  colorlinks=true,
  linkcolor=arxivlink,
  citecolor=arxivlink,
  urlcolor=arxivlink,
  pdftitle={AMPBench-MT: A Homology-Controlled Benchmark for Antimicrobial Peptide Potency, Spectrum, and Safety Prediction},
  pdfauthor={Ziheng Zhou, Huiyu Luo, Xiaohu Zhu, Nan Wang, Xuebiao Qin, Chaoyan Zhang, Jun Yan},
  pdfkeywords={AI4Science, antimicrobial peptides, benchmark, MIC regression, activity spectrum, toxicity prediction, selectivity, protein language models}
}

\excludecomment{CCSXML}
\providecommand{\Description}[1]{}
\newcommand{\ccsdesc}[2][]{\par\noindent\textbf{CCS Concepts}\enspace\textbullet\ \begingroup\def~{\enspace$\rightarrow$\enspace}#2.\endgroup\par}

\renewcommand{\headeright}{arXiv Preprint}
\renewcommand{\undertitle}{Preprint}
\renewcommand{\shorttitle}{AMPBench-MT}

\title{AMPBench-MT: A Homology-Controlled Benchmark for Antimicrobial Peptide Potency, Spectrum, and Safety Prediction}

\author[1]{Ziheng Zhou}
\author[2]{Huiyu Luo}
\author[3]{Xiaohu Zhu}
\author[4]{Nan Wang}
\author[1]{Xuebiao Qin}
\author[1]{Chaoyan Zhang}
\author[1]{Jun Yan\thanks{Corresponding author.}}

\affil[1]{Shanghai Ocean University, Shanghai, China}
\affil[2]{Tongji University, Shanghai, China}
\affil[3]{Center for Safe AGI, Shanghai, China}
\affil[4]{DP Technology, Shanghai, China}
\affil[ ]{\texttt{zihengzhouac@outlook.com} \quad \texttt{2211285@tongji.edu.cn} \quad \texttt{aleph@csagi.org} \quad \texttt{wangnan411570@gmail.com}}
\affil[ ]{\texttt{xbqin@shou.edu.cn} \quad \texttt{chyzhang@shou.edu.cn} \quad \texttt{yanjun@ieee.org}}

\date{}

\begin{document}
\maketitle

\begin{abstract}
Computational AMP discovery is often evaluated through AMP/non-AMP recognition, yet follow-up decisions depend on assay-derived evidence such as target-species potency, hemolysis, toxicity, and selectivity. Existing AMP and peptide benchmarks cover binary recognition, multilabel annotation, assay regression, or broader peptide-model comparison, but they do not jointly place AMP recognition, species-conditioned potency, spectrum, safety-facing proxy endpoints, and cross-endpoint behavior within one sequence-homology-controlled protocol. To address this problem, we introduce AMPBench-MT, a provenance-preserving benchmark that standardizes canonical peptide records and organizes them into binary recognition, species-conditioned pMIC regression, and endpoint-specific potency and safety-facing readouts. Across 161 endpoint-specific model evaluations, high binary performance does not reliably indicate assay-endpoint behavior. Frozen protein-language-model embeddings form the leading pMIC error cluster, while graph and classical regressors remain close. Spectrum labels further reveal that PR-oriented metrics can be misleading under scarce observed negatives, whereas low-toxicity, HC50 hemolysis, and selectivity expose smaller but more assay-facing signals. AMPBench-MT shows that AMP evaluation should move beyond recognition leaderboards toward endpoint-aware evidence auditing. Our proposed benchmark is available at \url{https://huggingface.co/datasets/ZihengZhou06/AMPBench-MT}.
\end{abstract}

\ccsdesc[500]{Computing methodologies~Machine learning}
\keywords{AI4Science, antimicrobial peptides, benchmark, MIC regression, activity spectrum, toxicity prediction, selectivity, protein language models}

\section{Introduction}

Machine learning has become a central computational route in AI-for-science research~\cite{wang2023scientific} and drug-discovery workflows~\cite{zhang2025artificial}, supporting molecular representation, compound design, preclinical prioritization, clinical development, and biomedical evidence synthesis. In preclinical discovery, these methods are used to search large candidate spaces while keeping activity, mechanism, and cytotoxicity evidence in view~\cite{catacutan2024machine}. Antibiotic-discovery work similarly uses model-guided search to prioritize candidates with interpretable activity evidence~\cite{wong2024discovery}. This shift makes evaluation design part of the discovery problem itself: candidate value is not determined by recognition alone, but by endpoint evidence that connects activity with safety-proxy readouts and developability constraints~\cite{pognan2023evolving}. For antimicrobial peptides (AMPs), the need is amplified by antimicrobial resistance, which remains a major global health burden~\cite{naghavi2024global}. This data-driven setting is especially relevant for AMPs because peptide sequence space is large, experimental assays are costly, and activity annotations are scattered across heterogeneous databases. Models can help triage candidates before synthesis or follow-up testing, but only if their predictions reflect the endpoints that guide prioritization~\cite{huang2023identification}. Recent AMP machine-learning reviews therefore emphasize not only identification and design, but also potency, spectrum, toxicity, hemolysis, selectivity, and related developability constraints~\cite{wan2024machine}.

Existing AMP and peptide benchmarks cover neighboring but still partial pieces of this evaluation landscape. Binary AMP benchmarks expose negative-data sensitivity~\cite{sidorczuk2022benchmarks}. ESCAPE organizes AMP activity as a multilabel benchmark~\cite{ojeda2026standardized}. QMAP covers MIC and HC50 assay regression~\cite{lavertu2026qmap}, while PepBenchmark broadens evaluation to peptide-machine-learning protocols~\cite{zhang2026pepbenchmark}. Table~\ref{tab:benchmark-comparison} summarizes this boundary. However, these benchmarks do not provide a single AMP-specific protocol that keeps binary recognition, species-conditioned potency, activity spectrum, toxicity, hemolysis, selectivity, and cross-endpoint prediction visible under the same controlled split and metric design.

\begin{table}[!t]
  \centering
  \small
  \caption{Benchmark-level comparison with closely related AMP and peptide benchmarks.}
  \label{tab:benchmark-comparison}
  \setlength{\tabcolsep}{2.2pt}
  \renewcommand{\arraystretch}{1.10}
  \begin{tabular}{@{}>{\raggedright\arraybackslash}p{0.23\columnwidth}>{\raggedright\arraybackslash}p{0.19\columnwidth}>{\raggedright\arraybackslash}p{0.27\columnwidth}>{\raggedright\arraybackslash}p{0.23\columnwidth}@{}}
    \toprule
    Work & Main focus & Coverage / protocol & Boundary vs. AMPBench-MT \\
    \midrule
    AMPBenchmark / negative-data bias study~\cite{sidorczuk2022benchmarks} &
    Binary AMP bias &
    AMP/non-AMP labels; negative-data sensitivity &
    No assay-level potency, spectrum, toxicity, hemolysis, or selectivity endpoints \\
    \addlinespace[1pt]
    ESCAPE~\cite{ojeda2026standardized} &
    Multilabel AMP classification &
    Activity/function labels with standardized annotation organization &
    Does not connect species-conditioned MIC, HC50, selectivity, and safety-proxy readouts \\
    \addlinespace[1pt]
    QMAP~\cite{lavertu2026qmap} &
    Assay-level AMP regression &
    MIC potency and HC50 hemolysis with homology-aware test sets &
    Closest assay comparator, but narrower than the AMPBench-MT endpoint set \\
    \addlinespace[1pt]
    PepBenchmark~\cite{zhang2026pepbenchmark} &
    General peptide ML benchmark &
    Standardized peptide datasets, preprocessing, model-family evaluation &
    Peptide-wide comparability rather than AMP-specific potency--safety-proxy linkage \\
    \addlinespace[1pt]
    AMPBench-MT &
    Endpoint-aware AMP prioritization &
    Binary AMP, species-conditioned pMIC, spectrum, low toxicity, HC50, selectivity, multitask; 30\% identity splits &
    Integrates recognition, potency, spectrum, and safety-proxy readouts under one AMP-specific protocol \\
    \bottomrule
  \end{tabular}
\end{table}

\begin{figure*}[!t]
  \centering
  \includegraphics[width=0.90\textwidth]{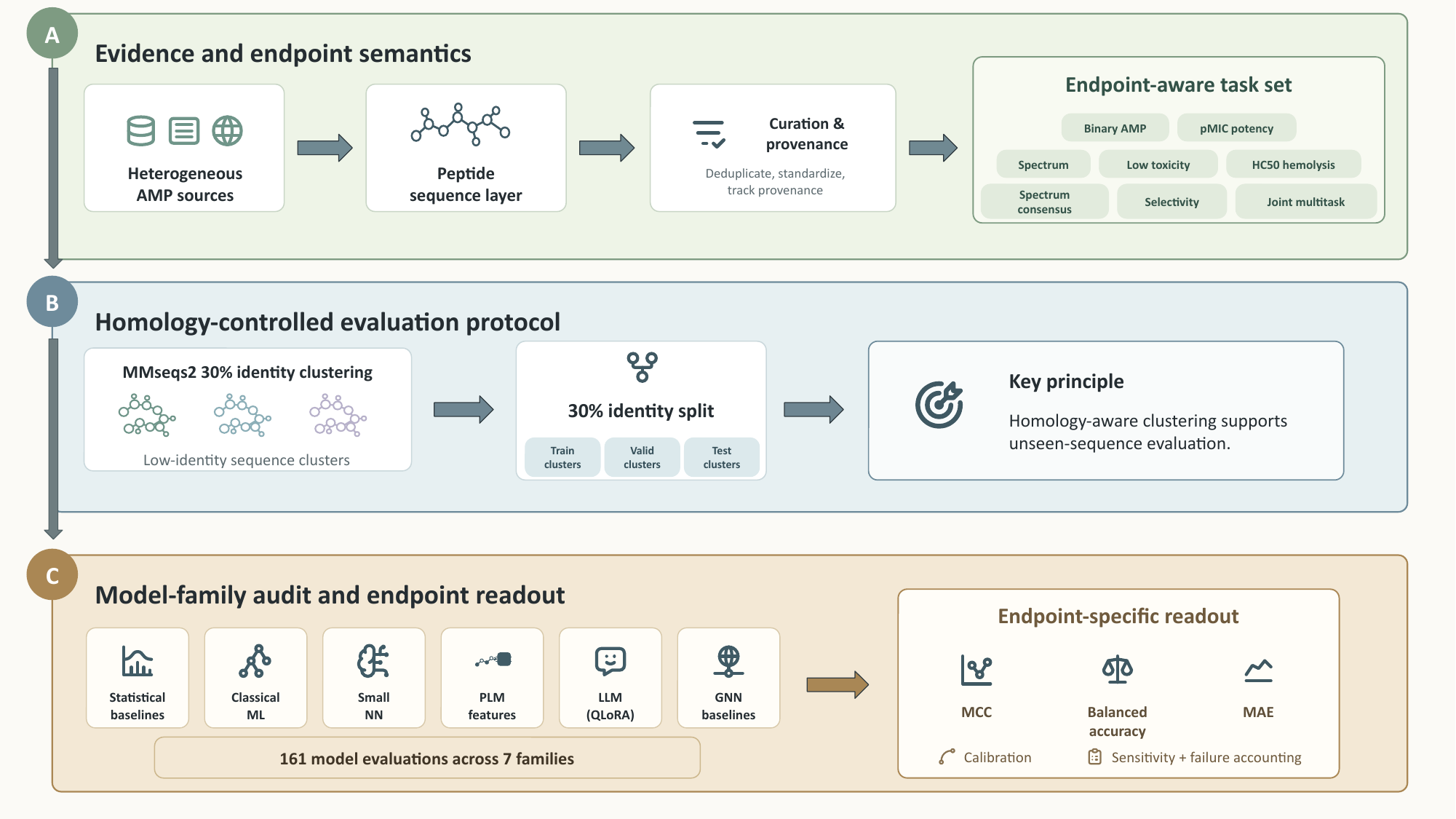}
  \caption{AMPBench-MT overview: multi-source evidence, homology-controlled splitting, and endpoint-specific evaluation.}
  \Description{A three-panel overview of evidence sources, MMseqs2 cluster splitting, and endpoint-level model evaluation.}
  \label{fig:ampbench-overview}
\end{figure*}

To fill this gap, we propose AMPBench-MT for probing peptide prioritization through assay-derived endpoint evidence: a model that recognizes AMP-like sequences can still fail on potency, spectrum, toxicity, hemolysis, or selectivity. AMPBench-MT standardizes peptide sequences, preserves endpoint provenance, applies MMseqs2 30\% cluster splits, and evaluates balanced AMP classification, species-conditioned MIC regression, multi-endpoint prediction, and joint pilot baselines. From a data-mining perspective, AMP benchmarking is an integration and evaluation problem: near-identical peptide backbones recur across databases, labels are endpoint-specific, missing assay evidence is not negative evidence, and class imbalance can make ranking metrics misleading. AMPBench-MT makes these failure modes measurable rather than hidden behind a single leaderboard. Figure~\ref{fig:ampbench-overview} gives the pipeline overview, while later sections provide the data-source, label-construction, split, and model-family details. Rather than reducing these settings to one aggregate leaderboard, the experiments use binary recognition as a historical anchor, MIC as a species-conditioned potency task, and the endpoint panel as a test of how label evidence and metric choice shape interpretation. AMPBench-MT controls near-duplicate peptide-sequence leakage, while reporting residual source, species, and publication overlap as benchmark boundaries rather than claiming prospective external validation.

Our contributions are summarized as follows:

\begin{itemize}
  \item \textbf{Benchmark construction.} AMPBench-MT provides a provenance-aware, homology-controlled AMP benchmark spanning binary recognition, species-conditioned MIC potency, activity spectrum, low toxicity, HC50 hemolysis, selectivity, and joint multitask evaluation.

  \item \textbf{Experimental protocol.} The experiments connect balanced binary classification, species-conditioned MIC regression, and multi-endpoint prediction through 161 endpoint-specific model evaluations across seven model families using endpoint-matched metrics and failure accounting.

  \item \textbf{Endpoint-level audit.} Results indicate that high binary recognition is not sufficient evidence for potency or safety-proxy assay readouts. Under the homology-controlled split, MIC regression clusters around protein-language-model embeddings with graph and tree-based comparators close behind, spectrum labels expose metric sensitivity under positive-heavy evidence, and two early shared multitask baselines provide pilot evidence for endpoint-dependent sharing.
\end{itemize}

\section{Related Work}

Drug-discovery machine learning provides the broader evaluation setting for AMP benchmarking. AI-for-science reviews frame models as tools for representation learning and scientific evidence synthesis~\cite{wang2023scientific}, while drug-development reviews place screening and candidate prioritization within longer discovery workflows~\cite{zhang2025artificial}. Preclinical candidate search raises similar concerns~\cite{catacutan2024machine}. Explainable antibiotic-class discovery raises related concerns~\cite{wong2024discovery}. Safety-proxy assay evidence enters from investigative toxicology, where activity evidence must be weighed against toxicity and developability risk~\cite{pognan2023evolving}. For AMPs, these lines of work set a benchmark target that is narrower than general drug discovery but broader than sequence recognition: potency and safety-proxy readouts must remain visible during model comparison.

AMP prediction has often been organized as binary classification between AMP and non-AMP sequences. This setting remains useful because it gives a common comparison point across external AMP tools, classical baselines, neural sequence models, and newer language-model approaches. Its limitation is semantic rather than only technical: a single AMP label does not encode MIC potency, pathogen spectrum, host-cell toxicity, hemolytic activity, or selectivity, all of which are emphasized in recent AMP machine-learning reviews~\cite{wan2024machine}. The benchmark itself can also shape the apparent difficulty of the task. Negative-data selection has been shown to bias AMP prediction benchmarks, making high binary scores sensitive to how non-AMP examples are constructed~\cite{sidorczuk2022benchmarks}. AMPBench-MT therefore keeps a balanced binary task for historical alignment, but treats it as one experimental setting rather than as evidence of assay-endpoint performance for follow-up decisions.

Recent work has moved AMP and peptide evaluation beyond this binary frame in complementary ways. ESCAPE organizes AMP activity information as a standardized multilabel benchmark for functional and activity annotation~\cite{ojeda2026standardized}. QMAP focuses on two quantitative assay endpoints, MIC and HC50, with homology-aware predefined test sets~\cite{lavertu2026qmap}. PepBenchmark broadens the view further by providing a unified pipeline for peptide machine learning across preprocessing choices, task families, and model types~\cite{zhang2026pepbenchmark}. Table~\ref{tab:benchmark-comparison} makes the distinction explicit: ESCAPE emphasizes multilabel AMP annotation, QMAP emphasizes MIC and HC50 assay regression, and PepBenchmark emphasizes peptide-wide protocol comparability. AMPBench-MT is AMP-specific and jointly covers binary recognition, species-conditioned MIC, spectrum, toxicity, hemolysis, selectivity, and cross-endpoint prediction under unified split construction and endpoint-matched metrics.

This endpoint framing also determines how model families are compared. AMPBench-MT evaluates protein-language-model embedding baselines~\cite{lin2023evolutionary} and generic LLM baselines adapted with parameter-efficient fine-tuning~\cite{dettmers2023qlora}, but it does not treat language-model scale as evidence of endpoint understanding. These model families are relevant to AMP prediction only when the task definitions, split construction, and endpoint metrics are fixed. The evaluation therefore places statistical baselines, classical machine learning, small neural sequence and shared multitask models, graph neural baselines, LLM/QLoRA baselines, and AMP-specific external tools within the same endpoint protocol. This design matters for imbalanced endpoint labels: a single classification-accuracy number can look acceptable when the majority class dominates, even if the model separates observed negatives poorly. MCC and balanced accuracy provide complementary views when one class dominates~\cite{chicco2020advantages}.

These previous studies motivate AMPBench-MT as an endpoint-aware benchmark rather than a single-task AMP classifier benchmark. The design retains binary AMP recognition for comparability and evaluates whether model behavior transfers to assay-derived endpoints under homology-controlled splits with endpoint-matched metrics.

\section{Benchmark Pipeline Design}

AMPBench-MT is constructed through three linked stages. Source records are first standardized into canonical peptide, assay, species, and provenance fields and routed to task-specific tables. Endpoint labels are then derived from explicit assay evidence, with missing evidence left unassigned rather than imputed. Finally, sequence clusters are used to create homology-controlled splits and verify the overlap boundaries reported with the release. The following subsections describe data standardization, endpoint construction, and split construction in this order.

\subsection{Task Scope and Data Standardization}

\begin{figure*}[!t]
  \centering
  \includegraphics[width=0.84\textwidth]{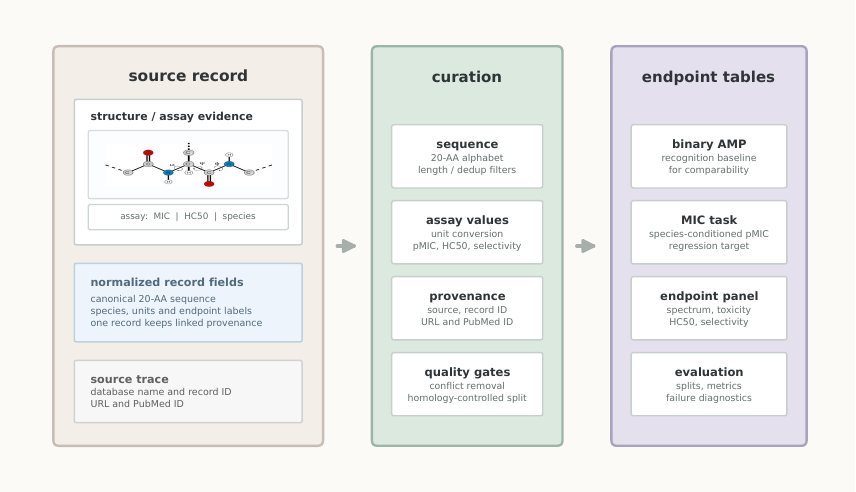}
  \caption{Record-level data curation and endpoint routing. Source records are standardized into canonical sequence, provenance, split, and endpoint tables before model evaluation.}
  \Description{A three-column schematic showing source records, curation steps, and endpoint tables for AMPBench-MT.}
  \label{fig:data-curation}
\end{figure*}

Figure~\ref{fig:data-curation} gives the record-level entry point to AMPBench-MT. Source entries may contain sequence, assay, activity, species, citation, and database identifiers. The pipeline standardizes these fields into canonical sequence and endpoint fields, applies quality gates, and routes usable records into binary, MIC, endpoint-panel, and evaluation tables. This record-first view matters because AMPBench-MT is organized around three settings chosen to separate historical AMP recognition, species-conditioned potency estimation, and downstream endpoint evidence. The binary setting retains AMP/non-AMP recognition as a historical anchor. The MIC setting pairs peptide sequences with target species so that potency is modeled as an assay-derived quantity rather than as a generic activity label. The multi-endpoint setting keeps spectrum, low toxicity, HC50 hemolysis, selectivity, and shared multitask prediction as separate evidence types. Main multi-endpoint tasks use uppercase 20-standard-amino-acid sequences of length 5--100, while the binary task keeps its historical 10--50 range. This defines a canonical sequence-only benchmark: it does not cover D-amino acids, terminal modifications, cyclization, non-canonical residues, amidation, salt forms, or related modified-peptide chemistry.

\begin{figure}[!t]
  \centering
  \includegraphics[width=0.94\textwidth]{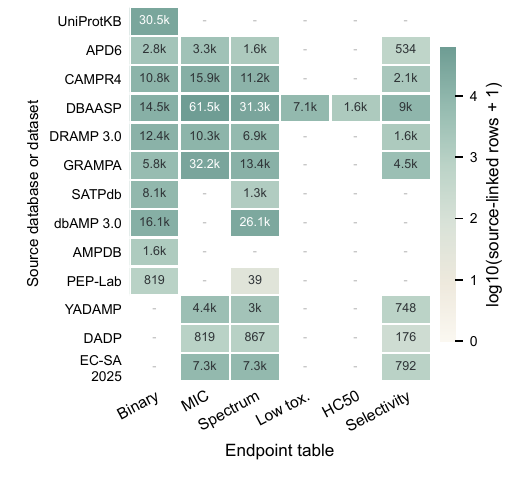}
  \caption{Source-linked endpoint coverage in the released benchmark tables. Cells report provenance-retaining endpoint rows with non-exclusive log-scaled counts.}
  \Description{A heatmap of source databases or datasets by endpoint table, showing multi-source coverage and DBAASP concentration for toxicity and HC50.}
  \label{fig:source-endpoint-coverage}
\end{figure}

The source layer is built from curated protein, AMP, bioactive-peptide, and assay databases rather than from a single repository. For binary recognition, the pipeline combines the Swiss-Prot subset of UniProtKB~\cite{uniprot2025uniprot} with AMP and bioactive-peptide repositories recorded in the provenance fields, including APD6~\cite{wang2026apd6}, CAMPR4~\cite{gawde2023campr4}, DBAASP~\cite{pirtskhalava2021dbaasp}, DRAMP 3.0~\cite{shi2022dramp}, GRAMPA~\cite{witten2019deep}, SATPdb~\cite{singh2016satpdb}, dbAMP 3.0~\cite{yao2025dbamp}, AMPDB~\cite{mondal2023developing}, and PEP-Lab~\cite{terziyski2023peplab}. Assay-bearing records for MIC and downstream endpoints are then drawn from DBAASP, CAMPR4, DRAMP 3.0, and GRAMPA. Additional MIC records come through Witten-derived exports~\cite{witten2019deep} spanning APD6, DBAASP, DRAMP 3.0, YADAMP~\cite{piotto2012yadamp}, and DADP~\cite{novkovic2012dadp}, together with the EC-SA 2025 AMP regression collection~\cite{cai2025bert}. Spectrum labels use target-group or activity annotations from DBAASP, CAMPR4, DRAMP 3.0, dbAMP 3.0, SATPdb, and PEP-Lab. Toxicity and HC50 labels are derived from DBAASP toxicity records. Across these tasks, the pipeline retains source database names, source IDs, URLs, PubMed IDs, and source-record links where available, so endpoint aggregation preserves provenance. Figure~\ref{fig:source-endpoint-coverage} summarizes this provenance layer and shows that endpoint coverage differs by source.

After standardization, each task table is defined by the observation unit required by its endpoint. The binary table is peptide-level, MIC regression is sequence--species level, spectrum depends on target-group or activity evidence, HC50 depends on exact hemolysis records, and selectivity depends on paired activity--hemolysis evidence. The resulting tables are therefore not forced to match by intersection or by peptide-level imputation. A peptide enters an endpoint only when the corresponding source evidence is present. Missing endpoint evidence remains missing rather than being converted into a negative, low-risk, or non-hemolytic label. This choice reduces label coverage, but it keeps task semantics aligned with the evidence needed for assay-aware evaluation. Task sizes and split allocations are therefore reported with the experimental protocol, where the consequences of endpoint-specific evidence are visible next to the model results.

\subsection{Endpoint Label Construction}

The pipeline keeps assay semantics explicit by converting heterogeneous assay records into endpoint-specific targets. For MIC regression, each input is a peptide sequence paired with a target species, and the target is the median pMIC of exact sequence--species evidence. We use MIC values reported in $\mu$M directly, scale nM and mM values to $\mu$M, and convert mass-concentration units such as $\mu$g/mL to $\mu$M using molecular weight. A direct-unit MIC sensitivity table excludes estimated mass-concentration conversions and is retained without train/validation/test partitioning. The reported MIC model results use the main split rather than a direct-unit-only evaluation. The pMIC target is defined as
\begin{align}
\mathrm{pMIC}
  &= -\log_{10}(\mathrm{MIC}_{\mathrm{mol/L}})
   = 6-\log_{10}(\mathrm{MIC}_{\mu\mathrm{M}}).
\label{eq:pmic}
\end{align}
Strict MIC pairs retain exact labels with at least one exact record, pMIC standard deviation at most 1.0 across repeated evidence, and median pMIC in the range 0--12. Repeated exact evidence is summarized by the median pMIC rather than by a single source record. The median aggregation prevents duplicated or repeated source records from being treated as independent labels, while still retaining exact sequence--species evidence when multiple records agree. It is a benchmark label rule rather than a meta-analysis of laboratory variability, so the resulting pMIC target should be interpreted as a standardized evaluation target. For classification endpoints, label construction is deliberately conservative: reported classification splits use rows with explicit endpoint labels, and absent activity evidence is not treated as a negative label. Exact MIC evidence is mapped into spectrum labels by
\begin{align}
y_{\mathrm{spectrum}}
  &=
  \begin{cases}
  1, & \mathrm{pMIC}\geq 5.0,\\
  0, & \mathrm{pMIC}\leq 4.0,\\
  \mathrm{excluded}, & 4.0<\mathrm{pMIC}<5.0.
  \end{cases}
\label{eq:spectrum-map}
\end{align}
Low-toxicity classification uses explicit toxicity labels, and HC50 uses exact hemolysis endpoints. Selectivity is computed from paired strict MIC and exact HC50 evidence as
\begin{align}
\mathrm{pHC50}
  &= -\log_{10}(\mathrm{HC50}_{\mathrm{mol/L}})
   = 6-\log_{10}(\mathrm{HC50}_{\mu\mathrm{M}}),\\
\mathrm{SI}_{\log} &= \mathrm{pMIC}-\mathrm{pHC50}.
\label{eq:selectivity}
\end{align}

Together, Eqs.~\eqref{eq:pmic}--\eqref{eq:selectivity} and the surrounding label rules prioritize traceable endpoint semantics over maximum label coverage. They also explain why the spectrum task must be read differently from the balanced binary task: an unobserved spectrum label is not treated as a confirmed negative, so AUPRC must be interpreted alongside MCC, balanced accuracy, and negative-class recall (specificity). Low-toxicity, HC50, and selectivity provide assay-derived safety-proxy evidence for follow-up decisions, but they are not clinical safety labels and should not be interpreted as evidence of in vivo safety.

\subsection{Homology-controlled Split Construction}

All main splits use sequence-cluster rather than row-level assignment because identical or near-identical peptides can recur across databases, assay records, species entries, and endpoint tables; row-level assignment could place the same backbone or a close homolog in both training and test through different provenance records. Using MMseqs2~\cite{steinegger2017mmseqs2}, the multi-endpoint protocol forms clusters at 30\% sequence identity with coverage 0.8, cov-mode 0, cluster-mode 2, and seed 42, and assigns each cluster to exactly one training, validation, or test partition. MIC follows the same cluster-level principle and additionally checks exact sequence--species pairs. The appendix overview in Table~\ref{tab:appendix-overview} reports the guarantees and boundary: all released tasks have complete row assignment and zero train--test overlap in exact peptide sequences or clusters formed at this threshold. However, the protocol does not hold out sources, databases, species, publications, or assays, and source databases may appear across splits. In MIC, 573 of 868 test species are represented in training, and 551 PubMed IDs are shared across the two partitions, covering 7,632 test rows. These counts describe provenance overlap, not reuse of exact sequences or sequence clusters. The joint multitask split globally assigns clusters formed at 30\% sequence identity using a greedy 70/15/15 split, so target-task test clusters cannot enter auxiliary-task training. Accordingly, this is a homology-controlled benchmark of unseen peptide clusters, not an external prospective validation study.

Three points carry into the experiments. First, provenance is retained before endpoint aggregation, so source identity and record links remain traceable after labels are routed into task tables. Second, endpoint tables are allowed to differ because MIC, spectrum, toxicity, HC50, and selectivity require different assay evidence. Third, all reported model comparisons use the same homology-controlled split logic, which keeps the emphasis on endpoint-specific behavior rather than on a single AMP-recognition score.

\section{Experiments}

The experiments mirror the three levels of the benchmark question. The binary task asks how models perform on AMP/non-AMP recognition under the benchmark split. MIC regression asks whether sequence and species information support quantitative potency prediction. The multi-endpoint and joint multitask setting asks whether models retain useful signal for spectrum, low toxicity, hemolysis, selectivity, and cross-endpoint sharing. Each subsection reports process, data, and test-set results within the homology-controlled setting. The goal is not to collapse these settings into one winner, but to show which readouts remain informative once endpoint semantics and split controls are fixed.

\subsection{Experimental Protocol and Metrics}
\label{sec:model-families}

The evaluation protocol is organized by endpoint rather than by a single score. Its 161 endpoint-specific model evaluations are grouped as historical anchors, representation probes, or stress tests rather than ranked as one leaderboard. They span seven model families: statistical baselines, classical ML with handcrafted peptide features, small neural sequence and shared multitask models, protein-LM embeddings, graph neural baselines, LLM/QLoRA baselines, and AMP-specific external tools. The inventory includes 38 binary AMP, 37 MIC, 19 spectrum, 14 spectrum-consensus, 19 low-toxicity, 17 HC50, 15 selectivity, and two early joint multitask evaluations. Classification tasks use AUROC, AUPRC, F1, MCC, balanced accuracy, and calibration-oriented scores when available. Thresholded metrics use validation-selected thresholds where recorded. Regression tasks use MAE, RMSE, $R^2$, Pearson, Spearman, and ranking accuracy when available. This metric design keeps binary recognition, potency estimation, spectrum prediction, safety-proxy labels, and multitask behavior as separate readouts. Repeated-seed gaps keep the results at the level of endpoint test-set evidence rather than statistical-significance claims.

Models discussed in the main-text rows carry citations there. Appendix-only PLM coverage adds ESM3-open-small and Ankh~\cite{hayes2025simulating,elnaggar2023ankh}; ProtBert shares the ProtTrans source already cited. Residue-graph rows use GCN, GAT, and GIN~\cite{kipf2016semi,velivckovic2017graph,xu2018powerful}, with PepGB as a peptide-graph design reference~\cite{lei2024pepgb}. Additional LLM rows cover Qwen2.5-Instruct and Mistral 7B~\cite{qwen2024qwen25,jiang2023mistral}, Gemma 3 and Phi-3.5~\cite{gemmateam2025gemma3,abdin2024phi3}, and Phi-4-mini~\cite{microsoft2025phi4mini}. CatBoost and MIC XGBoost are cited with the main results; multi-endpoint tree rows use LightGBM~\cite{ke2017lightgbm}. PepLM-GNN and PepGB are design references rather than direct task reproductions; external tools are historical rather than contamination-free baselines, and PLM/LLM pretraining overlap is unaudited. Main-text rows are representative; appendix figures retain full metrics and lower-ranked rows. The public release provides endpoint tables, splits, schema, validation report, manifest, checksums, and source-database licensing and terms metadata.

\begin{figure*}[!t]
  \centering
  \includegraphics[width=0.88\textwidth]{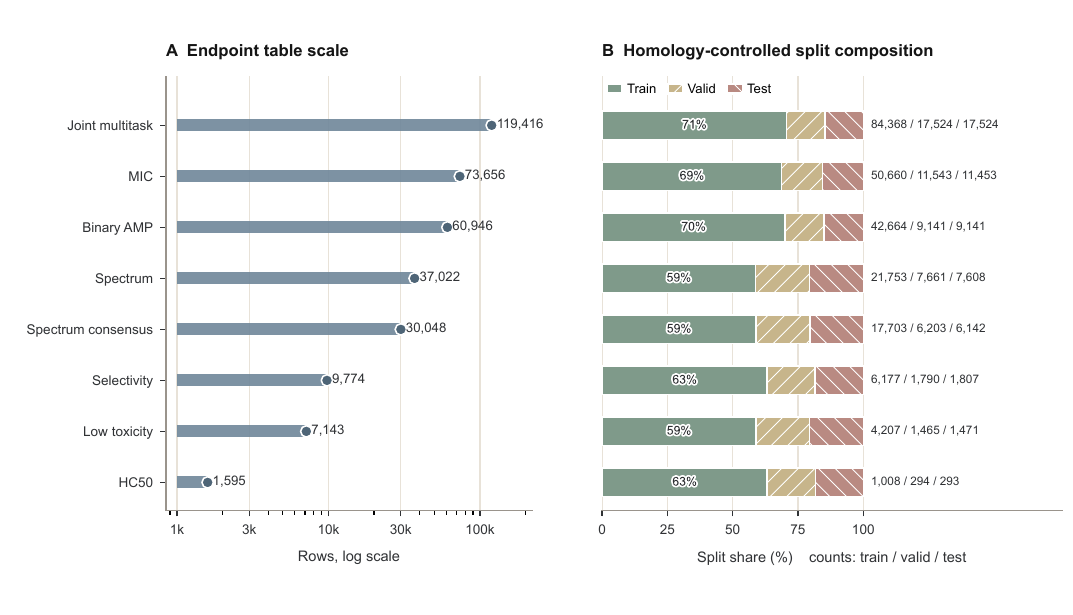}
  \caption{Endpoint-table sizes and homology-controlled train/validation/test allocation. The joint multitask total pools MIC, spectrum, low toxicity, and HC50 rows.}
  \Description{Panel A shows endpoint-table row counts on a log scale. Panel B shows train, validation, and test split shares with exact counts.}
  \label{fig:dataset-summary}
\end{figure*}

Figure~\ref{fig:dataset-summary} reports task sizes and split allocations as evidence availability rather than endpoint prevalence. MIC and joint multitask tables are large because they pool many exact sequence--species assay rows or multiple endpoint labels, whereas HC50 and selectivity require exact hemolysis or paired activity--hemolysis evidence. The low-toxicity table is the explicit-toxicity subset, and the direct-unit MIC sensitivity table is a unit-conversion diagnostic, not a model-evaluation split. These imbalances condition how AUPRC, MCC, regression error, and multitask sharing should be interpreted.

\subsection{Balanced AMP Binary Classification}

\textbf{Process.} The binary experiment evaluates model performance on historical AMP/non-AMP recognition under the homology-controlled split. The comparison includes AMP-specific external tools, protein-language-model embeddings with classifiers, and LLM/QLoRA baselines. Because the task is deliberately balanced, it is used as a historical anchor rather than as the primary evidence for assay-aware endpoint performance. Table~\ref{tab:binary-results} keeps the main text focused on MCC and balanced accuracy, while Fig.~\ref{fig:appendix-binary} reports the full binary metric set.

\textbf{Data.} The balanced task contains 60,946 peptides split into 42,664 training rows, 9,141 validation rows, and 9,141 test rows. It uses the historical 10--50 amino-acid range, whereas endpoint tasks retain 5--100 amino acids to preserve assay evidence. Cross-task conclusions are therefore endpoint-level rather than length-controlled model comparisons.

\begin{table}[!t]
  \centering
  \small
  \caption{Representative balanced AMP binary-classification readouts. Higher is better.}
  \label{tab:binary-results}
  \setlength{\tabcolsep}{2.4pt}
  \renewcommand{\arraystretch}{1.08}
  \begin{tabular}{@{}>{\raggedright\arraybackslash}p{0.60\columnwidth}rr@{}}
    \toprule
    Model & MCC & Bal. Acc. \\
    \midrule
    Qwen3-4B (QLoRA)~\cite{yang2025qwen3} & 0.862 & 0.930 \\
    ESM-1b 650M (PLM embed.)~\cite{rives2021biological} & 0.856 & 0.927 \\
    DeepSeek-R1-Distill-Llama-8B (QLoRA)~\cite{guo2025deepseek,grattafiori2024llama} & 0.853 & 0.926 \\
    Qwen3-8B (QLoRA)~\cite{yang2025qwen3} & 0.843 & 0.921 \\
    ProtGPT2 (PLM embed.)~\cite{ferruz2022protgpt2} & 0.839 & 0.919 \\
    ProtT5-XL-BFD (PLM embed.)~\cite{elnaggar2021prottrans} & 0.828 & 0.914 \\
    ampir (external tool)~\cite{fingerhut2020ampir} & 0.768 & 0.883 \\
    AMPScanner v2 (external tool)~\cite{veltri2018deep} & 0.695 & 0.846 \\
    \bottomrule
  \end{tabular}
  \vspace{1pt}
  \begin{minipage}{0.98\columnwidth}
    \small
    \emph{Note.} Rows are representative binary evaluations. Full metrics are in the appendix. External tools and pretrained models are historical references, not contamination-free baselines.
  \end{minipage}
\end{table}

\textbf{Results.} Binary recognition is strong across the leading binary rows. Qwen3-4B~\cite{yang2025qwen3} with QLoRA~\cite{dettmers2023qlora} reaches MCC 0.862 and balanced accuracy 0.930. A protein-language-model embedding baseline, ESM-1b 650M~\cite{rives2021biological}, is close with MCC 0.856, and the leading AMP-specific external tool, ampir~\cite{fingerhut2020ampir}, reaches MCC 0.768. These strong binary scores provide a historical recognition reference, but they do not establish species-conditioned potency, toxicity, hemolysis, spectrum, or selectivity performance.

\subsection{Species-conditioned MIC Regression}

\textbf{Process.} The MIC experiment evaluates quantitative potency as a sequence--species regression task. Rows provide peptide sequence and target species. Species encoding is summarized in the appendix overview. Figure~\ref{fig:mic-cluster} visualizes representative readouts from protein-language-model embeddings, graph neural baselines, classical feature-based regressors, small neural baselines, LLM/QLoRA numeric-generation baselines, and a statistical mean-backoff baseline. Fig.~\ref{fig:appendix-mic-classification} reports the complete MIC result set.

\textbf{Data.} The MIC task contains 73,656 exact sequence--species pairs, split into 50,660 training rows, 11,543 validation rows, and 11,453 test rows. Species identity is part of the prediction problem, so species-conditioned features are allowed and potency is evaluated conditioned on both peptide sequence and target organism. This is a species-conditioned homology split, not a species-held-out or source-held-out benchmark. Test species represented in training and publications shared across splits are reported as residual provenance risks rather than treated as eliminated leakage. The target is pMIC, where a one-unit difference corresponds to a ten-fold concentration difference on the MIC scale. Lower MAE and RMSE indicate more accurate potency estimates, while Pearson, Spearman, $R^2$, and ranking accuracy measure linear association, rank association, explained variance, and pairwise ordering quality.

\begin{figure}[!t]
  \centering
  \includegraphics[width=0.94\textwidth]{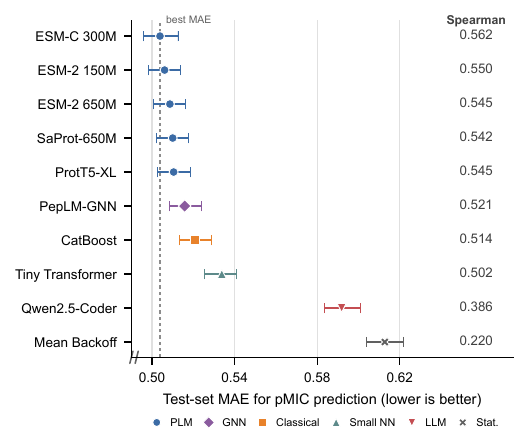}
  \caption{MIC regression comparison on the homology-controlled pMIC test split. Points show MAE with 95\% bootstrap intervals from 1000 resamples of retained single-run test predictions ($n=11{,}453$); Spearman is listed at right. The x-axis is zoomed from 0.49, lower MAE is better, and intervals are not multi-seed estimates.}
  \Description{A horizontal point-range plot comparing ten MIC regression rows by test MAE, with Spearman values displayed in a right-side column.}
  \label{fig:mic-cluster}
\end{figure}

\textbf{Results.} Protein-language-model embeddings form the leading MIC error cluster in the homology-controlled MIC evaluation. ESM-C 300M~\cite{evolutionaryscale2024esmcambrian} gives the lowest MAE at 0.504, with RMSE 0.667, Pearson 0.550, Spearman 0.562, $R^2$ 0.286, and ranking accuracy 0.700. The leading alternatives are tightly clustered: ESM-2 variants~\cite{lin2023evolutionary}, SaProt-650M~\cite{su2024saprot}, and ProtT5-XL-UniRef50~\cite{elnaggar2021prottrans} all fall between MAE 0.506 and 0.511. Figure~\ref{fig:mic-cluster} gives the same representative ordering with retained-prediction bootstrap intervals, showing that the leading PLM rows overlap under this single-run prediction resampling view. The PepLM-GNN graph row~\cite{yan2026peplm} follows this PLM cluster at MAE 0.516 and Spearman 0.521, numerically ahead of CatBoost on both readouts but still below the leading embeddings. Because this row adapts a peptide graph design to MIC regression rather than reproducing the original peptide--protein interaction task, the comparison is reported as a topology-aware baseline within the same split. The additional GNN rows in Fig.~\ref{fig:appendix-mic-classification} are less competitive; simple residue graphs did not match the strongest sequence-representation rows. Classical models remain competitive but slightly behind the leading embeddings, with CatBoost~\cite{prokhorenkova2018catboost} and XGBoost~\cite{chen2016xgboost} at MAE 0.521 and 0.522. The remaining small sequence neural baselines are weaker, and the Qwen2.5-Coder numeric-generation baseline~\cite{hui2024qwen2} with QLoRA~\cite{dettmers2023qlora} trails the embedding and classical models. The MIC task places PLM embeddings in the leading cluster, with the graph-based comparator adding a competitive but non-leading point of comparison. The margins among leading PLM variants remain modest rather than decisive. The best $R^2$ remains below 0.3, so the result should be read as lower error and stronger endpoint ranking within the benchmark rather than as solved quantitative potency prediction.

\subsection{Multi-endpoint and Joint Multitask Prediction}

\textbf{Process.} The multi-endpoint experiment asks whether models retain endpoint signal beyond binary AMP recognition and MIC potency. Single-task models cover spectrum, low toxicity, HC50 hemolysis, and selectivity. Two shared-encoder neural models cover joint MIC, spectrum, low toxicity, and HC50 prediction. Table~\ref{tab:multiendpoint-results} reports selected endpoint readouts rather than a full endpoint leaderboard.

\textbf{Data.} The single-task multi-endpoint tables include 37,022 spectrum rows, 30,048 spectrum-consensus rows, 7,143 low-toxicity rows, 1,595 HC50 rows, and 9,774 selectivity rows. The joint multitask table contains 119,416 rows across MIC, spectrum, low toxicity, and HC50, using the same low-toxicity subset. These counts are not harmonized into a complete endpoint matrix. They reflect the available evidence for each assay-derived or annotation-derived label. Strict spectrum maps pMIC $\geq 5$ (MIC $\leq 10~\mu$M) to positives and pMIC $\leq 4$ (MIC $\geq 100~\mu$M) to observed negatives, excluding the intervening decade. Low-toxicity retains rows with DBAASP-derived explicit toxicity labels. Label 1 means low toxicity, label 0 means observed toxicity, and missing toxicity evidence is not imputed. The spectrum tasks are positive-heavy evidence audits rather than ordinary balanced classifiers. Conventional supervised metrics are still reported, so absent target-group evidence is an interpretation boundary rather than redesigned training signal. MCC and balanced accuracy are necessary complements to AUPRC and F1. The safety-proxy readouts are smaller but more connected to follow-up decisions, which is why HC50 and selectivity are retained even though their regression tables are much narrower than MIC.

\begin{table}[!t]
  \centering
  \small
  \caption{Representative endpoint signals and failure modes. Complete endpoint results are shown in the appendix data figures.}
  \label{tab:multiendpoint-results}
  \setlength{\tabcolsep}{1.8pt}
  \renewcommand{\arraystretch}{1.02}
  \begin{tabular}{@{}>{\raggedright\arraybackslash}p{0.25\columnwidth}>{\raggedright\arraybackslash}p{0.36\columnwidth}>{\raggedright\arraybackslash}p{0.31\columnwidth}@{}}
    \toprule
    Endpoint & Representative readout & Observed pattern \\
    \midrule
    Spectrum strict & CatBoost~\cite{prokhorenkova2018catboost}: AUPRC 0.999; MCC 0.000 & High AUPRC; zero MCC \\
    Spectrum consensus & CatBoost~\cite{prokhorenkova2018catboost}: AUPRC 0.999; MCC 0.000 & Same metric pattern \\
    Low toxicity & ProtT5-XL-UniRef50~\cite{elnaggar2021prottrans}: AUPRC 0.823; MCC 0.534 & Nonzero MCC \\
    HC50 hemolysis & ESM-2 650M~\cite{lin2023evolutionary}: MAE 0.433; Spearman 0.553 & Positive rank association \\
    Selectivity & ESM-C 600M~\cite{evolutionaryscale2024esmcambrian}: MAE 0.621; $R^2$ 0.077 & Low explained variance \\
    Joint multitask & Shared Tiny Transformer multi-head~\cite{vaswani2017attention}: MIC MAE 0.561; HC50 MAE 0.499; spectrum MCC 0.000 & Mixed endpoint changes \\
    \bottomrule
  \end{tabular}
  \vspace{1pt}
  \begin{minipage}{0.98\columnwidth}
    \small
    \emph{Note.} Rows are representative of stronger readouts, diagnostic failures, or pilot multitask behavior. Appendix data figures retain the full model inventory and lower-ranked cases.
  \end{minipage}
\end{table}

\textbf{Results.} The multi-endpoint results differ by endpoint. Spectrum is the most pronounced metric-failure case: CatBoost~\cite{prokhorenkova2018catboost} reaches AUROC/AUPRC 0.810/0.999 in the strict spectrum setting, but MCC and balanced accuracy remain 0.000 and 0.500. The consensus-spectrum sensitivity row has the same pattern. This means that high PR-oriented scores do not imply useful observed-negative discrimination in the positive-heavy spectrum labels. Thresholded and calibration-oriented metrics are needed to expose whether the model is separating the scarce observed negatives. The appendix threshold diagnostic in Table~\ref{tab:appendix-overview} makes this failure mode explicit: CatBoost predicts all strict and consensus-spectrum test rows as positive at the validation-selected threshold, leaving no true negatives. The current spectrum endpoint should therefore be read mainly as a positive-evidence audit with limited observed-negative evidence. Low-toxicity classification is more informative at thresholded metrics: ProtT5-XL-UniRef50~\cite{elnaggar2021prottrans} reaches AUROC/AUPRC 0.826/0.823 and MCC/balanced accuracy 0.534/0.767. HC50 hemolysis regression also contains measurable signal. ESM-2 650M~\cite{lin2023evolutionary} reaches MAE/RMSE 0.433/0.587 and Spearman/$R^2$ 0.553/0.309. Selectivity is harder: ESM-C 600M~\cite{evolutionaryscale2024esmcambrian} gives the lowest selectivity MAE in the table at 0.621, but its $R^2$ is only 0.077. Joint multitask learning remains pilot evidence from two early shared baselines. The shared Tiny Transformer multi-head model~\cite{vaswani2017attention} improves over the shared CNN multi-head baseline on MIC MAE and HC50 MAE, but it does not resolve spectrum discrimination, where MCC and balanced accuracy remain 0.000 and 0.500. Fig.~\ref{fig:appendix-mic-classification} provides the complete endpoint rows. These results support endpoint-by-endpoint evaluation rather than assuming that binary recognition or joint sharing transfers uniformly to all assay targets.

\subsection{Experiment-level Summary}

Read together, the experiments separate three kinds of evidence. Binary AMP recognition is comparatively tractable and remains useful as a historical anchor, but the limited paired-model diagnostic reported in the appendix (Table~\ref{tab:appendix-overview}) does not support treating binary MCC as a proxy for MIC or spectrum behavior. MIC results point to a modest PLM-led potency cluster, with PepLM-GNN and classical regressors close but non-leading. The endpoint panel then shows where metric choice and label evidence matter most: spectrum can look strong under PR-oriented metrics while failing MCC and balanced accuracy, whereas low toxicity, HC50, and selectivity expose smaller but more assay-facing signals.

These patterns argue for endpoint-level evidence rather than a single leaderboard. Main-text figures and tables show representative high-performing, historically relevant, or diagnostic cases. The appendix retains full metrics, lower-ranked rows, and diagnostic notes. Broader claims about external assay generalization, model-family dominance, or reliable cross-endpoint transfer require separate validation.

\section{Conclusions and Future Work}

AMPBench-MT evaluates endpoint-aware prediction signals for AMP prioritization by combining provenance-aware labels, MMseqs2 30\% identity cluster splits, endpoint-specific metrics, and 161 endpoint-specific model evaluations across binary classification, species-conditioned MIC regression, and multi-endpoint prediction. The results keep binary recognition, species-conditioned potency, and behavior on safety-proxy readouts as separate evidence types: PLM embeddings form a modest leading MIC cluster, spectrum labels expose observed-negative discrimination limits, and early shared multitask baselines show mixed endpoint-dependent behavior. The evaluation controls sequence homology but does not hold out sources, databases, species, publications, or assays. Other limitations include limited uncertainty estimates, imbalanced spectrum labels, historical external tool baselines, unaudited overlap with PLM/LLM pretraining data, coverage restricted to canonical sequences, and a narrow multitask set. Future work should add external assay splits, held-out source, database, species, and assay diagnostics, multi-seed uncertainty, observed-negative spectrum evidence, modified-peptide coverage, and broader multitask architectures.

\bibliographystyle{ACM-Reference-Format}
\bibliography{references}

\clearpage
\appendix

\begin{figure}[H]
  \centering
  \includegraphics[width=\textwidth]{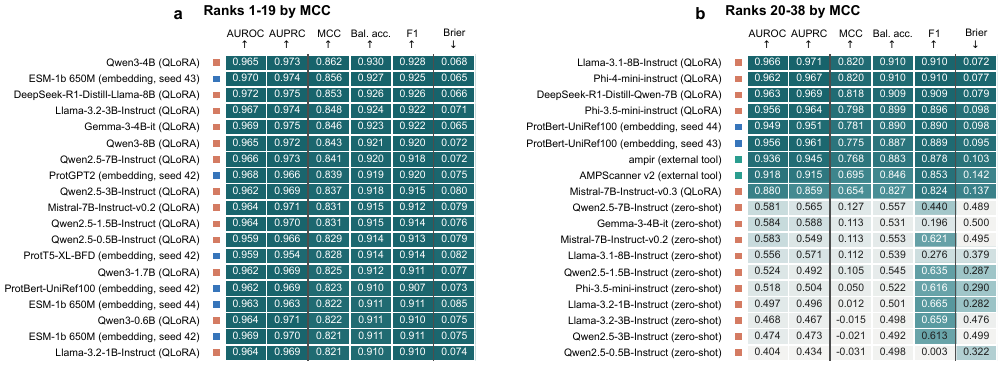}
  \caption{Complete binary AMP/non-AMP classification results. All 38 reported model rows are retained and ordered by MCC, with panels split only for readability. Each cell prints the exact test metric; arrows indicate the preferred direction. Family markers are blue for PLMs, coral for LLMs, and teal for external AMP tools. Higher AUROC, AUPRC, MCC, balanced accuracy, and F1 are better, whereas lower Brier score is better.}
  \label{fig:appendix-binary}
\end{figure}

\begin{figure}[H]
  \centering
  \includegraphics[width=0.96\textwidth]{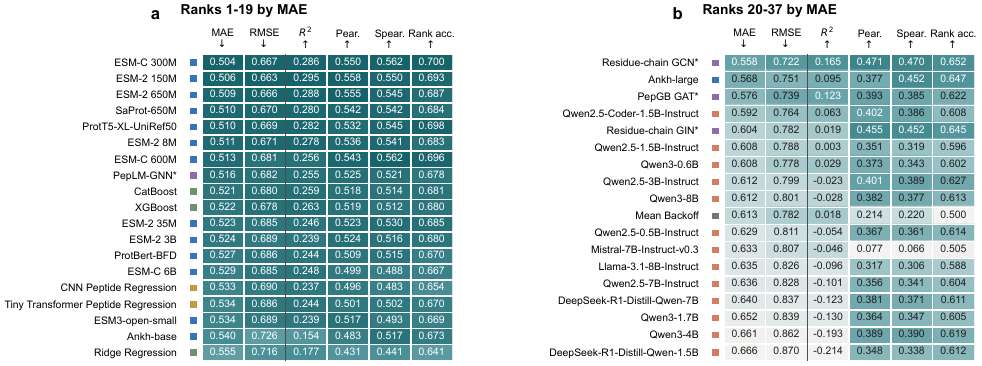}
  \par\vspace{-2pt}
  \includegraphics[width=0.96\textwidth]{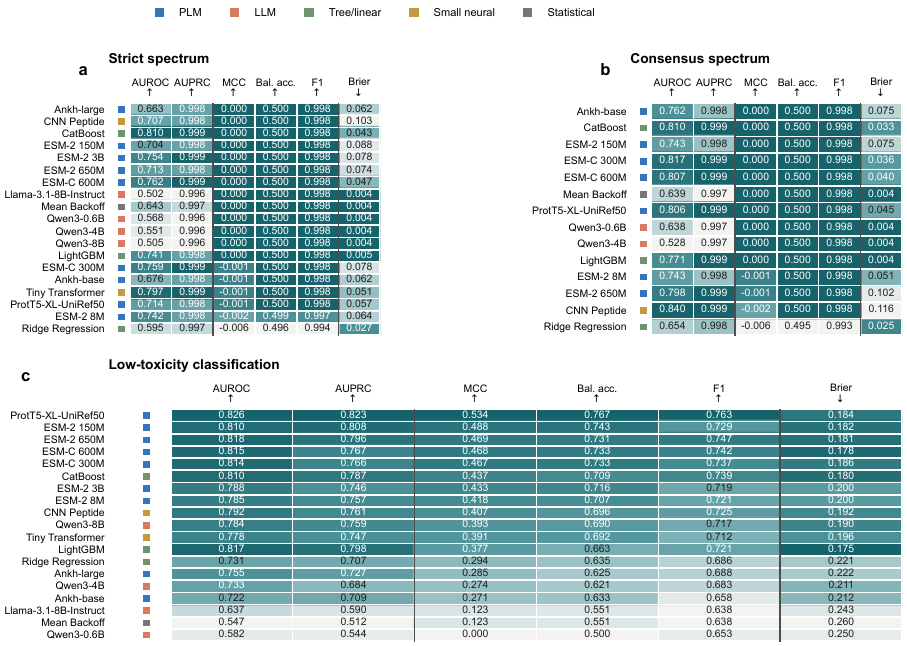}
  \caption{Complete species-conditioned MIC, spectrum, and low-toxicity results. The upper matrix retains all 37 MIC rows ordered by MAE; its family markers are blue (PLM), coral (LLM), purple (GNN), green (tree/linear), ochre (small neural), and gray (statistical). The lower matrix retains all reported strict-spectrum, consensus-spectrum, and low-toxicity rows and includes its family key. Cells print exact test metrics and arrows indicate the preferred direction. For MIC, lower MAE/RMSE and higher $R^2$, correlations, and ranking accuracy are preferred. Asterisks mark GNN rows adapted from the named graph designs rather than direct reproductions of their original tasks. Spectrum is a positive-evidence audit, so PR-oriented metrics are shown alongside thresholded and calibration-oriented metrics.}
  \label{fig:appendix-mic-classification}
\end{figure}

\begin{figure}[H]
  \centering
  \includegraphics[width=\textwidth]{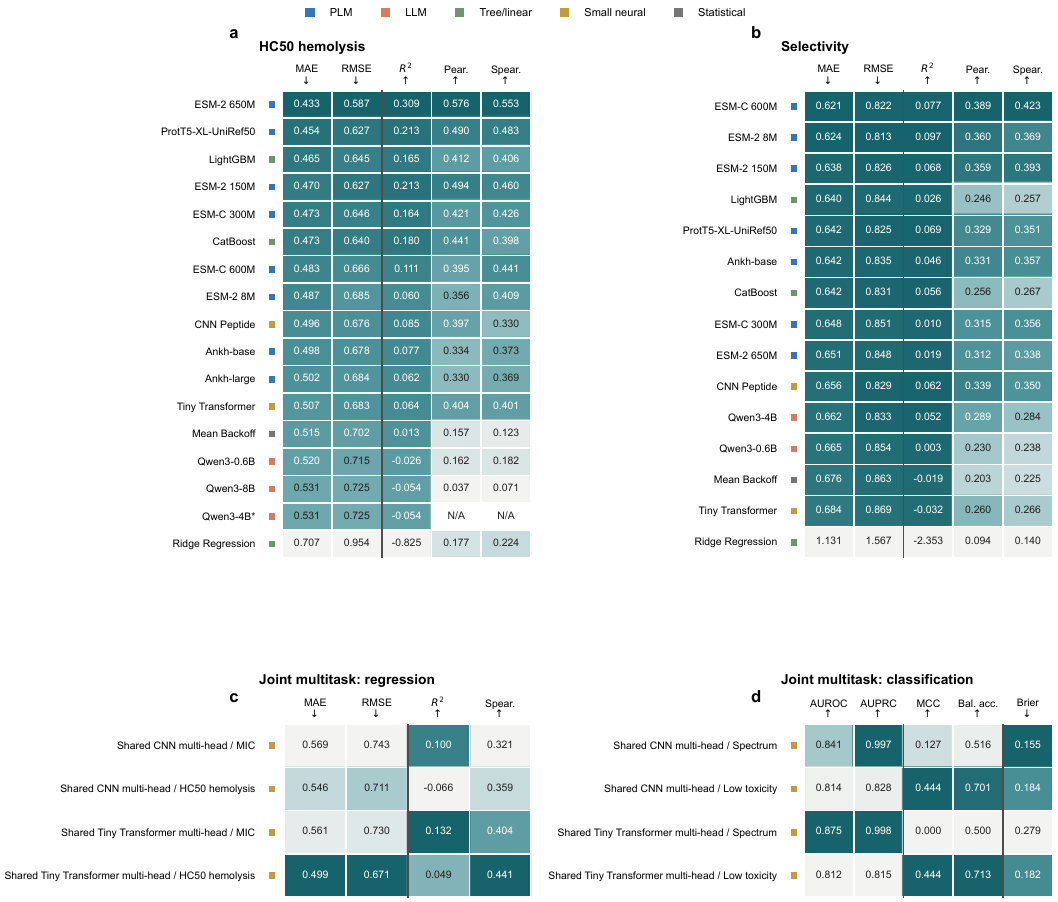}
  \caption{Complete HC50, selectivity, and joint multitask results. Panel (a) gives HC50 hemolysis regression, panel (b) gives selectivity regression, and panels (c,d) give the joint multitask regression and classification rows. All reported rows are retained for each endpoint. Each cell prints the exact test metric, arrows indicate the preferred direction, and the colored square identifies the model family. The asterisk marks the constant-prediction Qwen3-4B HC50 row; its Pearson and Spearman values are N/A because the prediction vector is constant. Other N/A cells indicate metrics not reported or not applicable to that endpoint.}
  \label{fig:appendix-safety}
\end{figure}

\noindent\textbf{Appendix.}
The appendix begins with the reproducibility and diagnostic overview in Table~\ref{tab:appendix-overview}. Complete model-level results are then shown as data figures: Fig.~\ref{fig:appendix-binary} covers binary AMP recognition, Fig.~\ref{fig:appendix-mic-classification} covers species-conditioned MIC regression, strict/consensus spectrum, and low-toxicity classification, and Fig.~\ref{fig:appendix-safety} covers HC50, selectivity, and joint multitask results. All reported model rows and metric values are retained in these figures; the cell shading is a within-metric visual aid and does not replace the printed values.

\begin{table}[H]
  \centering
  \small
  \caption{Appendix reproducibility, protocol, and diagnostic overview.}
  \label{tab:appendix-overview}
  \setlength{\tabcolsep}{2.2pt}
  \renewcommand{\arraystretch}{0.88}
  \begin{minipage}[t]{0.46\textwidth}
    \textbf{Evaluation inventory.}\par\vspace{1pt}
    \begin{tabular*}{\linewidth}{@{\extracolsep{\fill}}>{\raggedright\arraybackslash}p{0.72\linewidth}r@{}}
      \toprule
      Category & Evaluations \\
      \midrule
      Binary AMP & 38 \\
      MIC & 37 \\
      Spectrum & 19 \\
      Spectrum consensus & 14 \\
      Low toxicity & 19 \\
      HC50 hemolysis & 17 \\
      Selectivity & 15 \\
      Joint multitask & 2 \\
      \midrule
      Total endpoint-specific model evaluations & 161 \\
      \bottomrule
    \end{tabular*}
  \end{minipage}
  \hfill
  \begin{minipage}[t]{0.50\textwidth}
    \textbf{Classification label counts.}\par\vspace{1pt}
    \begin{tabular*}{\linewidth}{@{\extracolsep{\fill}}lrrrr@{}}
      \toprule
      Task & All + & All -- & Test + & Test -- \\
      \midrule
      Binary & 30,473 & 30,473 & 4,570 & 4,571 \\
      Spectrum & 36,625 & 397 & 7,578 & 30 \\
      Consensus & 29,651 & 397 & 6,118 & 24 \\
      Low tox. & 3,754 & 3,389 & 713 & 758 \\
      \bottomrule
    \end{tabular*}
  \end{minipage}

  \vspace{6pt}
  \textbf{Public release.}\par\vspace{1pt}
  \begin{tabular*}{\textwidth}{@{\extracolsep{\fill}}>{\raggedright\arraybackslash}p{0.20\textwidth}>{\raggedright\arraybackslash}p{0.75\textwidth}@{}}
    \toprule
    Item & Included files \\
    \midrule
    Public URL & \url{https://huggingface.co/datasets/ZihengZhou06/AMPBench-MT} \\
    Release & 2026-07-08 \\
    License & Research and Review Use License \\
    Data tables & task CSVs, split CSVs, complete tables (\texttt{all.csv}) \\
    Metadata/\allowbreak validation & schema, task overview, split summary, validation report \\
    Integrity & MANIFEST and SHA256 checksums \\
    \bottomrule
  \end{tabular*}

  \vspace{6pt}
  \textbf{Model protocol and MIC species encoding.}\par\vspace{1pt}
  \begin{tabular*}{\textwidth}{@{\extracolsep{\fill}}>{\raggedright\arraybackslash}p{0.16\textwidth}>{\raggedright\arraybackslash}p{0.28\textwidth}>{\raggedright\arraybackslash}p{0.27\textwidth}>{\raggedright\arraybackslash}p{0.22\textwidth}@{}}
    \toprule
    Family & Sequence/species input & Prediction head or evaluator & Protocol note \\
    \midrule
    Statistical & Train sequence mean, species mean, then global mean & Mean or backoff predictor & MIC reference row; no learned species channel \\
    Classical ML & Peptide features plus species one-hot, species mean, and species count priors & Linear, kernel, tree, or boosting heads & Unseen-species one-hot inputs are zero, with global or zero numeric priors \\
    Small neural & Amino-acid tokens plus train-fitted species embedding & CNN, MLP, Tiny Transformer, or multi-head shared encoder & Lightweight controlled baselines \\
    PLM embeddings & Frozen protein-LM sequence embedding plus train-fitted species embedding and numeric row features & Shallow classifier or regressor & MIC PLM heads map unseen species to a reserved unknown-species index \\
    GNN & Residue graph plus a training-split species index & GCN/GAT/GIN-style message passing & MIC-only single-seed comparators \\
    LLM/QLoRA & Prompt serializes peptide sequence and target species string & QLoRA adapter or zero-shot parser & Strict numeric parsing is applied, with failures reported as diagnostics \\
    External AMP tools & Publicly available sequence-only AMP-specific tools & Tool score or provided predictor & Binary historical baselines; no MIC species channel \\
    \bottomrule
  \end{tabular*}
\end{table}

\clearpage

\begin{table}[H]
  \centering
  \small
  \setlength{\tabcolsep}{2.2pt}
  \renewcommand{\arraystretch}{0.94}

  \textbf{Post-hoc diagnostic checks.}\par\vspace{1pt}
  \begin{tabular*}{\textwidth}{@{\extracolsep{\fill}}>{\raggedright\arraybackslash}p{0.15\textwidth}>{\raggedright\arraybackslash}p{0.20\textwidth}>{\raggedright\arraybackslash}p{0.42\textwidth}>{\raggedright\arraybackslash}p{0.16\textwidth}@{}}
    \toprule
    Check & Scope & Diagnostic result & Boundary \\
    \midrule
    Endpoint transfer & Models matched by reported name & MIC: $n=10$, Spearman $\rho=-0.04$ for binary MCC vs. $-\mathrm{MAE}$; spectrum: $n=4$ with endpoint MCC fixed at 0; low tox.: $n=4$, $\rho=0.60$ & Diagnostic only; small paired sets \\
    Split provenance & Released task splits & Sequence/cluster train--test overlap is 0 for all tasks; source databases are shared; 573/868 MIC test species occur in training, with 551 shared PubMed IDs & Residual-risk diagnostic; not held-out validation \\
    MIC unit sensitivity & Main vs. direct-unit MIC tables & Direct-unit MIC sensitivity table has 65,658 pairs; 65,640 are shared with the main MIC table & Sensitivity table only; no model-evaluation split \\
    Spectrum threshold & CatBoost strict/consensus & At threshold 0.05, all test rows are predicted positive: strict TP/FP/TN/FN = 7578/30/0/0; consensus = 6118/24/0/0 & Explains MCC/Bal. Acc. failure \\
    MIC bootstrap & 1000 test-set resamples & ESM-C 300M: 0.504 [0.496, 0.512]; PepLM-GNN: 0.516 [0.508, 0.524]; CatBoost: 0.521 [0.513, 0.529] & Single-run prediction bootstrap \\
    \bottomrule
  \end{tabular*}

  \vspace{1pt}
  \begin{minipage}{0.98\textwidth}
    \small
    \emph{Note.} The overview records the released task inventory, release contents, label counts, model-family input conventions, and residual-risk diagnostics. In the class-count block, + denotes label 1 and -- denotes label 0; for low toxicity, label 1 means low toxicity. Diagnostic rows use retained metric/prediction files, split files, or release files as indicated. Endpoint-transfer correlations match model rows by reported name across endpoints; they are not controlled paired experiments. Provenance rows are residual-risk diagnostics rather than held-out validation, and MIC intervals are single-run prediction bootstraps rather than multi-seed uncertainty. Complete model-level results are shown in Figs.~\ref{fig:appendix-binary}--\ref{fig:appendix-safety}.
  \end{minipage}
\end{table}

\clearpage

\end{document}